\documentclass[twocolumn,preprint]{elsarticle}

\usepackage{lineno,hyperref}
\usepackage{url}
\usepackage{hyperref}
\usepackage{times}
\usepackage{epsfig}
\usepackage{graphicx}
\usepackage{amsmath}
\usepackage{multirow}
\usepackage{amssymb}
\usepackage{subfig}
\usepackage{graphicx}
\usepackage{float}
\usepackage{booktabs}
\usepackage{textcomp}
\nolinenumbers

\journal{Neurocomputing}

\begin{document}

\twocolumn[{
\begin{frontmatter}

\title{CS-AF: A Cost-sensitive Multi-classifier Active Fusion Framework \\ for Skin Lesion Classification}

\author{Di Zhuang, Keyu Chen, and J. Morris Chang\\
{\tt\small \{dizhuang, keyu, chang5\}@usf.edu}\\}
\address{Department of Electrical Engineering, University of South Florida, Tampa, FL 33620}




\begin{abstract}
Convolutional neural networks (CNNs) have achieved the state-of-the-art performance in skin lesion analysis. Compared with single CNN classifier, combining the results of multiple classifiers via fusion approaches shows to be more effective and robust.
Since the skin lesion datasets are usually limited and statistically biased, while designing an effective fusion approach, it is important to consider not only the performance of each classifier on the training/validation dataset, but also the relative discriminative power (e.g., confidence) of each classifier regarding an individual sample in the testing phase,
which calls for an active fusion approach.
Furthermore, in skin lesion analysis, the data of certain classes (e.g., the benign lesions) is usually abundant making them an over-represented majority, while the data of some other classes (e.g., the cancerous lesions) is deficient, making them an underrepresented minority. It is more crucial to precisely identify the samples from an underrepresented (i.e., in terms of the amount of data) but more important minority class (e.g., certain cancerous lesion). In other words, misclassifying a more severe lesion to a benign or less severe lesion should have relative more cost (e.g., money, time and even lives).
To address such challenges, we present CS-AF, a cost-sensitive multi-classifier active fusion framework for skin lesion classification.
In the experimental evaluation, we prepared 96 base classifiers (of 12 CNN architectures) on the ISIC research datasets. Our experimental results show that our framework consistently outperforms the static fusion competitors. \end{abstract}

\begin{keyword}
\texttt{Deep Neural Networks; Multi-classifier Fusion; Active Fusion; Ensemble Learning; Cost-sensitive Classification; Skin Lesion Analysis.}
\end{keyword}
\end{frontmatter}
}]


\section{Introduction} \label{Introduction}
Deep learning (DL) has achieved great success in many applications related to skin lesion analysis. For instance, Zhang et al. \cite{zhang2019attention} has shown that convolutional neural networks (CNNs) have achieved the state-of-the-art performance in skin lesion classification. Also, as the development of various deep learning techniques, numerous different designs of classifiers, that might have different CNN architectures, use different sizes of the training data, use different subsets or classes distributions of the training data or use different feature sets, were proposed to tackle the skin lesion classification problem.
For instance, as shown in the ISIC Challenges \cite{gutman2016skin, codella2018skin, codella2019skin}, several CNN architectures have been used in skin lesion analysis, including ResNet, Inception, DenseNet, PNASNet, etc.
Because of such difference (i.e., CNN architectures, subset of the training data, feature sets, etc.), those classifiers tend to have distinct performance under different conditions (e.g,, different subsets or classes distributions of data). There is no one-size-fits-all solution to design a single classifier for skin lesion classification. It is necessary to investigate multi-classifier fusion techniques to perform skin lesion classification under different conditions.

Designing an effective multi-classifier fusion approach for skin lesion classification needs to address two challenges. 
First, since the datasets are usually limited and statistically biased \cite{gutman2016skin, codella2018skin, codella2019skin}, while conducting multi-classifier fusion, it is necessary to consider not only the performance of each classifier on the training/validation dataset, but also the relative discriminative power (e.g., confidence) of each classifier regarding an individual sample in the testing phase. This challenge requires the researchers to design an active fusion approach, that is capable of tuning the weight assigned to each classifier dynamically and adaptively, depending on the characteristics of given samples in the testing phase.
Second, since in most of the real-world skin lesion datasets \cite{gutman2016skin, codella2018skin, codella2019skin} 
the data of certain classes (e.g., the benign lesions) is abundant making them an over-represented majority, while the data of some other classes  (e.g., the cancerous lesions) is deficient, making them an underrepresented minority,
it is more crucial to precisely identify the samples from an underrepresented (i.e., in terms of the amount of data) but more important minority class (e.g., certain cancerous lesion).
For instance, a deadly cancerous lesion (e.g., melanoma) that rarely appears during the examinations should be barely misclassified as benign or other less severe lesions (e.g., dermatofibroma). Specifically, misclassifying a more severe lesion to a benign or less severe lesion should have relative more cost (e.g., money, time and even lives). Hence, it is also important to enable such ``cost-sensitive'' feature in the design of an effective multi-classifier fusion approach for skin lesion classification.

In this work, we propose CS-AF, a cost-sensitive multi-classifier active fusion framework for skin lesion classification, where we define two types of weights: the objective weights and the subjective weights.
The objective weights are designed according to the classifiers' reliability to recognize the particular skin lesions, which is determined by the prior knowledge obtained through the training phase.
The subjective weights are designed according to the relative confidence of the classifiers while recognizing a specific previously ``unseen'' sample (i.e., individuality), which are calculated by the posterior knowledge obtained through the testing phase. While designing the objective weights, we also utilize a customizable cost matrix to enable the ``cost-sensitive'' feature in our fusion framework, where given a sample, different outputs (i.e., the correct prediction or all kinds of errors) of a
classifier should result in different costs.
For instance, the cost of misclassifying melanoma as benign should be much higher than misclassifying benign as melanoma.
In the experimental evaluation, we trained 96 base classifiers as the input of our fusion framework, utilizing twelve CNN architectures, four classes distributions and two data split ratios on the ISIC research datasets for skin image analysis \cite{gutman2016skin, codella2018skin, codella2019skin}. Our experimental results show that our CS-AF framework consistently outperforms the static fusion competitors in terms of accuracy, and always achieves lower total cost.

To summarize, our work has the following contributions:

$\bullet$ We present a novel and effective cost-sensitive multi-classifier active fusion framework, CS-AF. To the best of our knowledge, this is the first work to apply active fusion for skin lesion analysis, and show its advantages over the conventional static fusion approaches.

$\bullet$ Our framework is the first to define the simple but effective objective/subjective weights and the customizable cost matrix, which enables the ``cost-sensitive'' feature for skin lesion analysis.

$\bullet$ A comprehensive experimental evaluation using twelve popular and effective CNN architectures has been conducted on the most popular skin lesion analysis benchmark dataset, ISIC research datasets \cite{gutman2016skin, codella2018skin, codella2019skin}. For the sake of reproducibility and convenience of future studies about fusion approaches in skin lesion analysis, we have released our prototype implementation of CS-AF, information regarding the experiment datasets and the code of our comparison experiments. \footnote[1]{\url{https://tinyurl.com/tabzxec}}

The rest of this paper is organized as follows:
Section~\ref{RelatedWork} presents the related literature review.
Section~\ref{Methodology} presents the notations of cost-sensitive active fusion, and describes our proposed framework.
Section~\ref{ExperimentalEvaluation} presents the experimental evaluation.
Section~\ref{Conclusion} concludes.

\section{Related Work} \label{RelatedWork}

\subsection{Multi-classifier Fusion} \label{RelatedWork_2}
Fusion approaches have been widely applied in numerous applications, such as skin lesion analysis \cite{perez2019solo, di2020saia}, human activity recognition \cite{tao2014ensemble, zhuang2020utility}, active authentication \cite{wu2016cost}, facial recognition \cite{ding2017trunk, nguyen2019autogan, zhuang2017fripal}, botnet detection \cite{mai2018cluster, zhuang2017peerhunter, zhuang2018enhanced} and community detection \cite{tagarelli2017ensemble, zhuang2019dynamo}. According to whether the weights are dynamically/adaptively assigned to each classifier, the multi-classifier fusion approaches are divided into two categories: (i) static fusion, where the weight assigned to each participating classifier will be fixed after its initial assignment, and (ii) active fusion, where the weights are adaptively tuned depending on the characteristics of given samples in the testing phase. Many conventional approaches, such as the bagging \cite{breiman1996bagging}, boosting \cite{schapire1990strength, freund1995desicion} and stacking \cite{wolpert1992stacked}, are static fusion. 
To date, a few methods attempting to conduct active fusion were also proposed \cite{ren2018multi, cruz2015meta, garcia2018dynamic}. For instance,
Ren et al. \cite{ren2018multi} proposes to use the decision credibility that is evaluated by fuzzy set theory and cloud model, to determine the real-time weight of a base classifier regarding the current sample in the testing phase.
META-DES \cite{cruz2015meta} defines five distinct sets of meta-features to measure the level of competence of a classifier for the classification of input samples, and proposes to train a meta-classifier to determine the rank or weight of a base classifier while facing input samples.
DES-MI \cite{garcia2018dynamic} propose an active fusion approach where the weights are determined via emphasizing more on the classifiers that are more capable of classifying examples in the region of underrepresented area among the whole sample distribution. In our work, we propose a novel active fusion approach, that leverages the ``reliability'' (Section~\ref{Objective_w}) and the ``individuality '' (Section~\ref{Subjective_w}) of the base classifiers to assign the weights dynamically and adaptively. 

\subsection{Fusion of CNNs for Skin Lesion Analysis} \label{RelatedWork_1}
Convolutional neural networks (CNNs) have achieved the state-of-the-art performance \cite{gutman2016skin, codella2018skin, codella2019skin} in skin lesion analysis since 2016 (i.e., ISIC 2016 Challenge \cite{gutman2016skin}), where nearly all the teams employed CNNs in either feature extraction or classification procedure. Recently, approaches attempting to apply fusion on CNNs to tackle the skin lesion classification, are proposed \cite{marchetti2018results, bi2017automatic, perez2019solo}.
For instance,
Marchetti et al. \cite{marchetti2018results} presents a fusion of CNNs framework for the the classification of melanomas versus nevi or lentigines, where five fusion approaches were implemented to fuse 25 different CNN classifiers trained on the same dataset of the same problem to make a single decision.
Bi et al. \cite{bi2017automatic} proposes another CNNs fusion framework to tackle the classification of melanomas versus seborrheic keratosis versus nevi, where three ResNet classifiers were trained for three different classification problems via fine-tuning pretrained ImageNet CNNs: the original three-class problem and two binary classifiers (i.e., melanoma versus both other lesion classes and seborrheic carcinoma versus both other lesion classes).
Perez et al. \cite{perez2019solo} conducts a comparison study between two fusion strategies for melanoma classification: selecting the classifiers at random (i.e., among 125 models over 9 CNNs architectures), and selecting the classifiers depending on their performance on a validation dataset.
To summarize, most of the existing approaches use static fusion for skin lesion analysis. However, as discussed in Section~\ref{Introduction}, since the skin lesion datasets are usually limited and statistically biased \cite{gutman2016skin, codella2018skin, codella2019skin}, it is necessary to enable active fusion in such problem. To the best of our knowledge, our work is the first to design and apply active fusion approach in skin lesion classification.

\subsection{Cost-sensitive Machine Learning} \label{RelatedWork_3}
A variety of cost-sensitive machine learning approaches have been proposed to tackle the class imbalance issue in pattern classification and learning problems. Mollineda et al. \cite{mollineda2007class}, a comprehensive study on the class imbalance issue, divides most of the cost-sensitive machine learning approaches into two categories: the data-level and the algorithmic-level. The data-level approaches usually manipulate the data distribution via over-sampling the samples of the minority classes or under-sampling the samples of the majority classes. For instance, SMOTE \cite{chawla2002smote} is an over-sampling technique proposed to address the over-fitting problem via synthesizing more of the samples of the minority classes. Several variants of the SMOTE approach \cite{wang2014hybrid, han2005borderline, bunkhumpornpat2009safe, maciejewski2011local} are also proposed to solve this issue. The algorithmic-level approaches directly re-design the machine learning algorithms to minimize a customizable loss function, that enables the ``cost-sensitive'' feature, of the classifier
on certain dataset (e.g., improving the sensitivity of the classifier towards minority classes). For instance,
Zhang et al. \cite{zhang2016evolutionary} proposes an extreme learning machine (ELM) based evolutionary cost-sensitive classification approach, where the cost matrix would be automatically identified given a specific task (i.e., which error cost more).
Iranmehr et al. \cite{iranmehr2019cost} extends the standard loss function of support vector machine (SVM) to consider both the class imbalance (i.e., the cost) and the classification loss.
Khan et al. \cite{khan2017cost} proposes a cost-sensitive deep neural network framework that could automatically learn the ``cost-sensitive'' feature representations for both the majority and minority classes, where during the training phase, the proposed framework would perform a joint-optimization on the class-dependent costs and the deep neural network parameters.
In this work, we enable the ``cost-sensitive'' feature in the process of multi-classifier fusion, and employ it in the skin lesion classification problem.

\begin{figure*}[!h]
  \centering
  \includegraphics[width=0.8\linewidth]{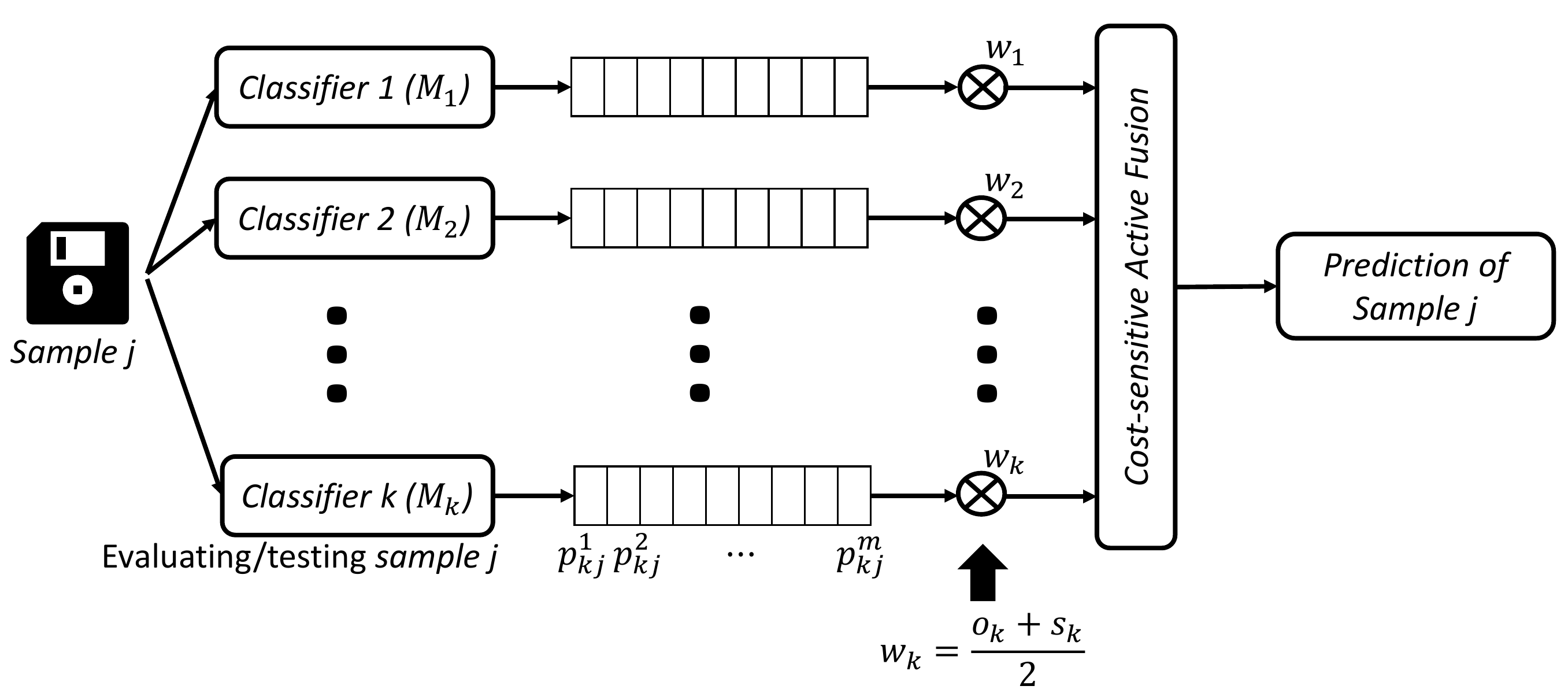}
  \caption{The Overview of CS-AF Framework.}
  \label{Overview}
\end{figure*}

\section{Methodology} \label{Methodology}
\subsection{Multi-classifier Fusion} \label{ES_Problem}
In multi-classifier fusion, we define a classification space, as shown in Figure~\ref{Overview}, where there are $m$ classes and $k$ classifiers. Let $\mathcal{M}=\{M_{1}, M_{2}, \dots, M_{k}\}$ denote the set of base classifiers and $\mathcal{C}=\{C_{1}, C_{2}, \dots, C_{m}\}$ denote the set of classes. Let $p^{m}_{kj}$ denote the posterior probability of given sample $j$ identified by classifier $M_{k}$ as belonging to class $C_{m}$, where $P_{kj}=\{p^{1}_{kj}, p^{2}_{kj}, \dots, p^{m}_{kj}\}$ and $\sum_{l=1}^{m} p^{l}_{kj} = 1$. Hence, all the posterior probabilities form a $k \times m$ decision matrix as follows:

\begin{equation}
P_{j} =
\begin{bmatrix}
p^{1}_{1j} &  p^{2}_{1j} & \cdots & p^{m}_{1j}\\
p^{1}_{2j} &  p^{2}_{2j} & \cdots & p^{m}_{2j}\\
\vdots &  \vdots & \ddots & \vdots\\
p^{1}_{kj} &  p^{2}_{kj} & \cdots & p^{m}_{kj}\\
\end{bmatrix}
\label{decision_matrix}
\end{equation}

Since the importance of different classifiers might be different, we assign a wight $w_{i}$ to the decision vector (i.e., posterior probabilities vector) of each classifier $C_{i}$, where $i \in \{1, 2, \dots, k\}$. Let $P_{m}(j)$ denote the sum of the posterior probabilities, that sample $j$ belonging to class $m$, of all the classifiers. Then, we have

\begin{equation}
P_{m}(j)=\sum_{i=1}^{k} w_{i} \cdot p^{m}_{ij}
\label{final_posterior_probability}
\end{equation}

The final decision (i.e., class) $D(j)$ of sample $j$ is determined by the maximum posterior probabilities sum:

\begin{equation}
D(j) = \textit{$\underset{i}{max} \ P_{i}(j)$}, \ i \in \{1, 2, \dots, m\}
\label{final_decision}
\end{equation}

Conventional multi-classifier fusion approaches either use the same weight for all the classifiers (i.e., average fusion) or use static weights that will not be changed after its initial assignment during the training phase. As illustrated in Figure~\ref{Overview}, our weights (i.e., 
$w_{k}=\frac{O_{k} + S_{k}}{2}$) contains two components:
(i) the objective weight $O_{k}$ that is static and determined by the prior knowledge obtained through the training phase (Section~\ref{Objective_w}), and (ii) the subjective weight $S_{k}$ that is dynamic and calculated by the posterior knowledge obtained through the testing phase (Section~\ref{Subjective_w}). To be simplified, we assign the same weight, i.e., $0.5$, on both $O_{k}$ and $S_{k}$, while combining them together.


\begin{figure*}[!h]
  \centering
  \includegraphics[width=0.9\linewidth]{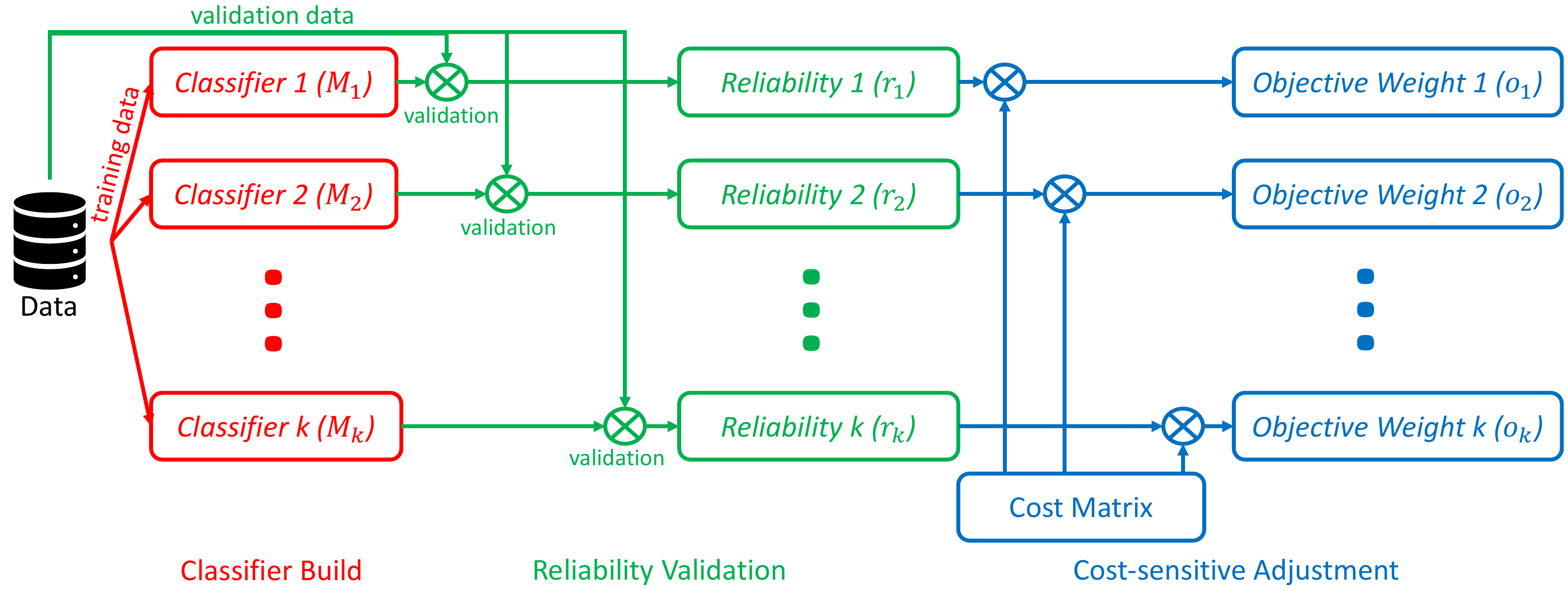}
  \caption{The calculation of objective weights.}
  \label{objective}
\end{figure*}

\begin{figure*}[!h]
  \centering
  \includegraphics[width=0.9\linewidth]{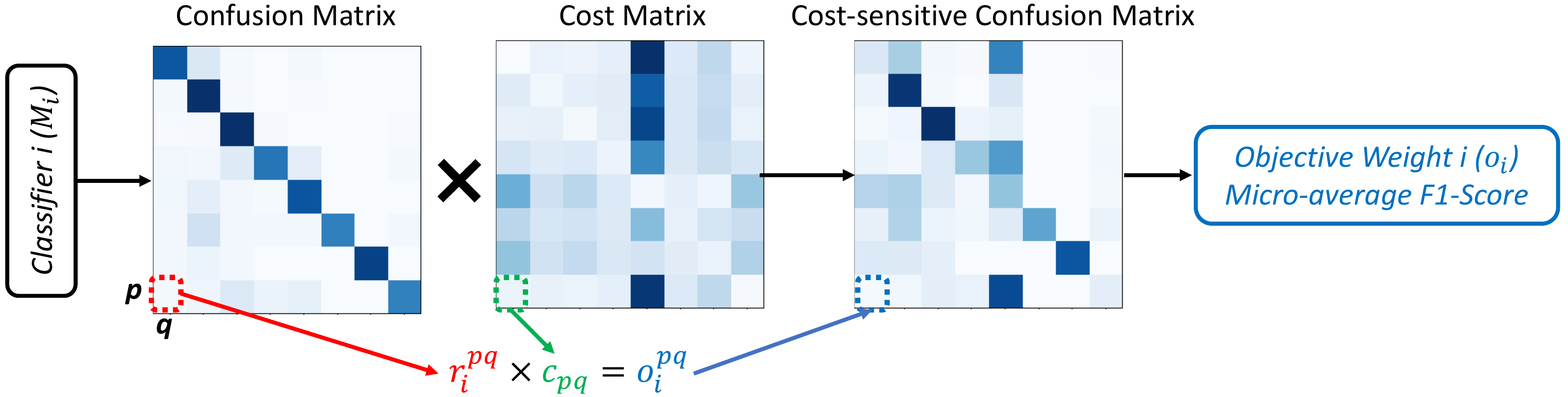}
  \caption{The calculation of cost-sensitive confusion matrix.}
  \label{Cost_CP}
\end{figure*}

\subsection{Cost-sensitive Problem Formulation} \label{Cost_Problem}
As discussed in Section~\ref{Introduction}, given a sample, different outputs (i.e., the correct prediction or all kinds of errors) of a classifier should result in different costs. For instance, misclassifying a more severe lesion to a benign or less severe lesion should have relative higher cost. Let $c_{pq}$ denote the cost of classifying an instance belonging to class $p$ into class $q$. Then, we obtain a cost matrix as follows:
\begin{equation}
CM =
\begin{bmatrix}
c_{11} &  c_{12} & \cdots & c_{1m}\\
c_{21} &  c_{22} & \cdots & c_{2m}\\
\vdots &  \vdots & \ddots & \vdots\\
c_{m1} &  c_{m2} & \cdots & c_{mm}\\
\end{bmatrix}
\label{cost_matrix}
\end{equation}

Let $W=\{w_{1}, w_{2}, \dots, w_{k}\}$ be a fusion weight vector, and $\mathcal{W}$ be the fusion weight vector space, where $W \in \mathcal{W}$. The goal of cost-sensitive multi-classifier fusion is to find the $W^{*} \in \mathcal{W}$, that can minimize the average cost of the fusion approach's outcomes over all the testing samples.
In Section~\ref{CM_Design}, we provide certain principles to design a good cost matrix. We also provide two examples of cost matrices, that emphasize on different skin lesion classes, based on our literature references and those principles.

\begin{figure*}[!h]
  \centering
  \includegraphics[width=0.9\linewidth]{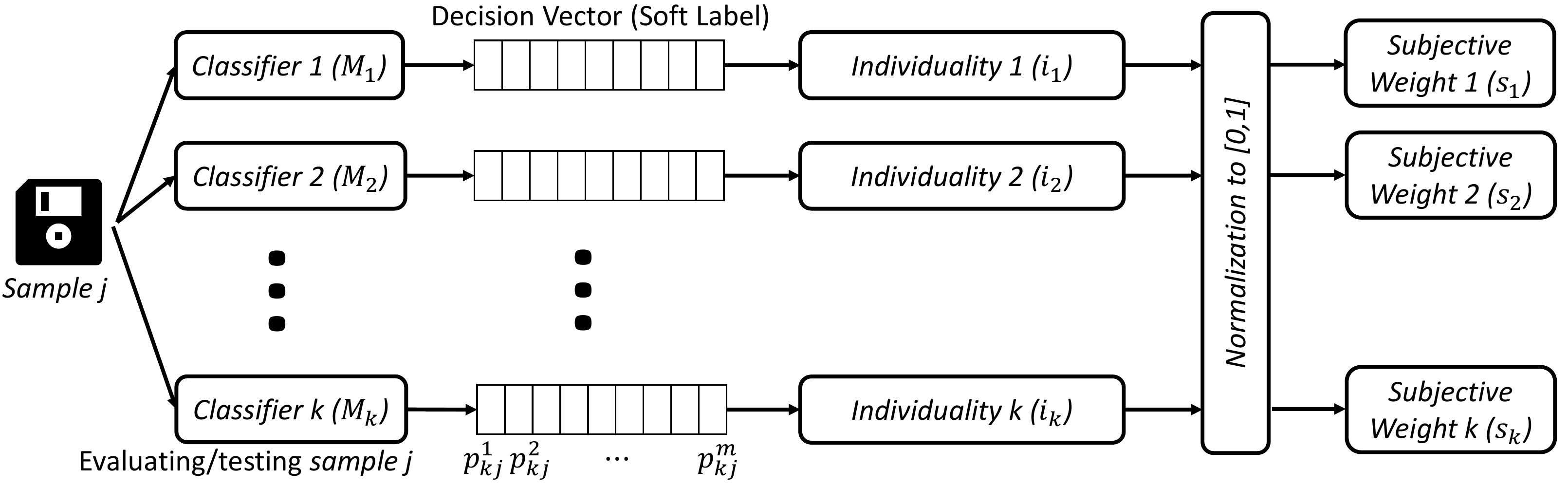}
  \caption{The calculation of subjective weights.}
  \label{subjective}
\end{figure*}

\subsection{Computing the Objective Weights} \label{Objective_w}
The objective weights are designed according to the classifiers’ reliability to recognize the particular skin lesions, which is determined by the prior knowledge obtained through the training phase.
In the training phase, we separate all the labelled data into two parts: training data and validation data. As shown in Figure~\ref{objective}, computing the objective weights in our framework contains three steps:

$\bullet$ Classifier build. We prepare a set of base classifiers, where all the classifiers might have different CNN architectures, use different size of the training data, or use different subset or classes distributions of the training data. In this step, we trained 96 base classifiers, more details are introduced in Section~\ref{Experiment_Design}.

$\bullet$ Reliability validation. Let $r_i$ denote the reliability of a base classifier $M_{i}$, that
is designed to describe the average recognition performance of the classifier on the validation data. Higher accuracy and less error on the validation data usually means higher reliability. Hence, we use the confusion matrix result of each base classifier on the same validation dataset as its reliability, where a confusion matrix \cite{visa2011confusion} is a table that is often used to describe the performance of a classifier on a set of validation data for which the true values are known. It allows easy identification of confusion between classes, e.g., one class is commonly mislabeled as the other. Many performance measures could be computed from the confusion matrix (e.g., F-scores). As such, we use $r_{i}^{pq}$ to denote the probability of a base classifier $M_{i}$ classifying an instance belonging to class p into class
q.

$\bullet$ Cost-sensitive adjustment. As described in Section~\ref{Cost_Problem}, we would like to enable the ``cost-sensitive'' feature in the design of our objective weights. As shown in Figure~\ref{Cost_CP}, for each classifier $M_{i}$, we use an element-wise multiplication between its reliability $r_{i}$ (confusion matrix) and the customized cost matrix (Section~\ref{Cost_Problem}) to formulate a cost-sensitive confusion matrix, where all the results/errors in the confusion matrix have been adjusted based on the cost matrix. Then, we use the micro-average F1-score \cite{van2013macro} of the cost-sensitive confusion matrix to define the objective weight of each base classifier, and all the object weights are automatically normalized to $(0, 1]$.

\subsection{Computing the Subjective Weights} \label{Subjective_w}
The subjective weights are designed according to the relative confidence of the classifiers while recognizing a specific previously “unseen” image (i.e., individuality), which are calculated by the posterior knowledge obtained through the testing phase.
As shown in Figure~\ref{subjective}, computing the subjective weights in our framework contains three steps:

$\bullet$ Sample evaluating/testing. Each sample is evaluated/tested through all the classifiers to obtain the corresponding decision vectors (i.e., the soft labels).

$\bullet$ Individuality calculation. We consider the individuality of a classifier as its relative class discriminative power regarding a given individual sample.
A classifier can easily identify the class of a given sample in the testing phase, if its posterior probabilities of the corresponding decision vector is highly concentrated in one class, and the misclassification rate would also be low. On the contrary, if the distribution of the posterior probabilities is close to uniform, the classifier shows its difficulty in discriminating the class of the given sample.  Also, different classifiers would present different distribution of the posterior probabilities in the decision vectors while testing the same sample. Hence, we define the the individuality $i_{k}$ of a classifier $M_{k}$ using the posterior probabilities distribution as follows:
    \begin{equation}
        i_{k}=\frac{1}{m-1}\sum^{m}_{l=1}(p^{*}_{kj}-p^{l}_{kj})
        \label{individuality}
    \end{equation}
where $p^{*}_{kj}$ is the largest posterior probability value in $P_{kj}$.

$\bullet$ Normalization. Since the subjective weights are relative values among all the base classifiers, we normalize each individuality $i_{k}$ to the subjective weight $S_{k} \in [0, 1]$ as follows:
    \begin{equation}
        S_{k}=\frac{i_{k}-i^{min}}{i^{max}-i^{min}}
        \label{normalization}
    \end{equation}
    where $i^{min}=\underset{j \in {1, 2, \dots, k}}{min} i_{j}$ and $i^{max}=\underset{j \in {1, 2, \dots, k}}{max} i_{j}$.

\section{Experimental Evaluation} \label{ExperimentalEvaluation}
We conducted our experiments on the ISIC Challenge 2019 dataset \cite{combalia2019bcn20000, tschandl2018ham10000, codella2018skin} and utilized 12 CNN architectures to evaluate the performance of our proposed CS-AF framework. Two examples of cost matrices, that emphasize on different skin lesion classes (i.e., cancerous lesion classes vs. benign lesion classes), have been designed to evaluate the effectiveness of the ``cost-sensitive'' feature of our proposed CS-AF framework.
Furthermore, extensive comparisons have been made among Max Voting Fusion (static), Average Fusion (static), AF (i.e., active fusion without the ``cost-sensitive'' feature) and our CS-AF framework. The presented results show that our approach consistently outperforms the base static fusion approaches (i.e., Max Voting and Average Fusion) in terms of the overall accuracy and the total cost, and consistently better than AF in terms of the total cost under different conditions.

\subsection{Experiment Dataset} \label{Experiment_Dataset}
In our experiments, we utilized the well known ISIC Challenge 2019 dataset \cite{combalia2019bcn20000, tschandl2018ham10000, codella2018skin}. Since the ground truth of the original testing data was not available, we only employed the original training data without meta-data in our experimental evaluation. This dataset (i.e., the original training data of the ISIC Challenge 2019) contains 25,331 images in total, coming from 3 source datasets: BCN\_20000 \cite{combalia2019bcn20000}, HAM10000 \cite{tschandl2018ham10000} and MSK \cite{codella2018skin}. It depicts 8 skin lesion diseases (i.e., 8 classes): melanoma (MEL, 4,522 images), melanocytic nevus (NV, 12,875 images), basal cell carcinoma (BCC, 3,323 images), actinic keratosis (AK, 867 images), benign keratosis (BKL, 2624 images), dermatofibroma (DF, 239 images), vascular lesion (VASC, 253 images) and squamous cell carcinoma (SCC, 628 images). We split the entire 25,331 images into training (80\%), validation (5\%) and testing (15\%) datasets.

\begin{table}
  \caption{The number (ratio) of samples of each skin lesion classes of different training datasets.}
  \label{data_distribution}
  \scalebox{0.67}{
  \begin{tabular}{ccccc}
    \toprule
    \textbf{Skin Lesion} & \textbf{Dist-1} & \textbf{Dist-2} & \textbf{Dist-3} & \textbf{Dist-4}\\
    \midrule
    MEL & 3,662 (18.1\%)  & 2,509 (12.4\%) & 5,052 (22.5\%) & 604 (2.7\%)  \\
    \midrule
    SCC & 502 (2.5\%)  & 2,510 (12.4\%) & 4,331 (19.3\%) & 1,200 (5.4\%) \\
    \midrule
    BCC & 2,670 (13.2\%) & 2,494 (12.4\%) & 3,781 (16.9\%) & 1,812 (8.1\%) \\
    \midrule
    NV & 10,235 (50.5\%) & 2,512 (12.4\%) & 3,032 (13.5\%)& 2,529 (11.3\%) \\
    \midrule
    AK & 705 (3.5\%)& 2,564 (12.5\%) & 2,463 (11.0\%)& 3,150 (14.1\%) \\
    \midrule
    DF & 188 (1\%)& 2,444 (12.3\%) & 1,871 (8.3\%) & 3,702 (16.9\%)\\
    \midrule
    VASC & 194 (1\%)& 2,522 (12.5\%) & 1,262 (5.6\%)& 4,334 (19.3\%) \\
    \midrule
    BKL & 2,099 (10.4\%)& 2,612 (13.0\%) & 626 (2.8\%)& 5,056 (22.6\%) \\
    \bottomrule
    \end{tabular}}
\end{table}

\begin{table*}
  \caption{The performance (accuracy in \%) of the base classifiers of twelve CNN architectures trained on 8 different training datasets.}
  \label{table1}
  \scalebox{0.84}{
  \begin{tabular}{ccccccccc}
    \toprule
    \textbf{CNN Architectures} & \textbf{Dist-1} & \textbf{Dist-1 Sub-70} & \textbf{Dist-2} & \textbf{Dist-2 Sub-70} & \textbf{Dist-3} & \textbf{Dist-3 Sub-70} & \textbf{Dist-4} & \textbf{Dist-4 Sub-70} \\
    \midrule
    PNASNet-5-Large \cite{liu2018progressive} & 78.48  & 76.53 & 81.14 & \textbf{80.73} & 77.01 & 75.21 & 78.76 & 75.34\\
    \midrule
    NASNet-A-Large \cite{zoph2016neural} & 78.35  & 76.71 & 79.80 & 78.31 & 76.00 & 75.36 & 76.12 & 74.32\\
    \midrule
    ResNeXt101-32$\times$16d \cite{xie2017aggregated} & 79.47 & 76.96 & \textbf{83.09} & 80.08 & \textbf{80.18} & \textbf{77.14} & \textbf{79.47} & \textbf{77.47}\\
    \midrule
    SENet154 \cite{hu2018squeeze} & \textbf{80.31} & \textbf{77.72} & 81.19 & 76.33 & 79.04 & 74.04 & 78.76 & 76.43\\
    \midrule
    Dual Path Net-107 \cite{chen2017dual} & 76.61 & 74.51 & 79.10 & 77.92 & 76.07 & 70.88 & 77.24 & 74.80\\
    \midrule
    Xception \cite{chollet2017xception} & 74.63 & 74.07 & 78.53 & 75.19 & 76.46 & 72.93 & 75.82 & 74.30\\
    \midrule
    Inception-V4 \cite{szegedy2017inception} & 76.76 & 74.22 & 80.11 & 77.45 & 77.09 & 75.37 & 75.89 & 74.10\\
    \midrule
    InceptionResNet-V2 \cite{szegedy2017inception} & 77.58 & 76.64 & 70.81 & 77.39 & 77.77 & 76.12 & 76.48 & 74.01\\
    \midrule
    SE-ResneXt101-32$\times$4d \cite{hu2018squeeze} & 77.45 & 77.21 & 79.87 & 78.41 & 75.38 & 74.68 & 75.60 & 74.33 \\
    \midrule
    ResNet152 \cite{he2016deep} & 75.69 & 73.23 & 79.27 & 75.01 & 76.00 & 74.96 & 75.77 & 74.45\\
    \midrule
    Inception-V3 \cite{szegedy2016rethinking} & 75.16 & 73.82 & 79.52 & 78.83 & 73.69 & 72.41 & 75.62 & 72.07\\
    \midrule
    EfficientNet-b7 \cite{tan2019efficientnet} & 67.31 & 63.07 & 74.10 & 71.28 & 71.81 & 68.75 & 71.48 & 67.37\\
    \bottomrule
    \end{tabular}}
\end{table*}

To evaluate the performance of our proposed CS-AF framework using the base classifiers that are trained from the datasets with different classes distributions, we designed 4 training datasets that have different classes distributions. For instance, one training dataset could have balanced classes distribution, and the other training datasets could have unbalanced classes distributions in different ways. The details (i.e., classes distributions) of each training dataset are shown in Table \ref{data_distribution} and described as below:
\begin{itemize}
    \item Dist-1: This training dataset follows the classes distribution of the original training dataset of the ISIC Challenge 2019 dataset.
    \item Dist-2: This training dataset contains evenly distributed number of samples for all classes.
    \item Dist-3: This training dataset contains more samples for cancerous lesion classes, and less samples for benign lesion classes.
    \item Dist-4: This training dataset contains less samples for cancerous lesion classes, and more samples for benign lesion classes (i.e., the opposite order of classes distributions as in Dist-3).
\end{itemize}

To generate different training datasets satisfying different classes distributions described above, we utilized data augmentation techniques to generate more images for the skin lesion classes lacking of images, and randomly sampled smaller portions from the classes with superfluous images. The main data augmentation techniques utilized are rotation (for 45, 90, 135, 180, 225, 270 and 315 degrees, respectively), horizontal flip or the combination of both. We also utilized the same strategy to generate the validation and testing datasets, to ensure the numbers of samples of all classes are equal, where there are approximate 200 samples of each class in the validation dataset, and approximate 500 samples of each class in the testing dataset.

In addition, to evaluate the performance of our proposed CS-AF framework using the base classifiers that are trained from different subsets of the training dataset, for each of those four training datasets that have different classes distributions, we randomly select 70\% of it to produce another sub-dataset, namely, Sub-70. Therefore, in our experimental evaluation, there are 8 different training datasets in total (i.e., Dist-1, Dist-1 Sub-70, Dist-2, Dist-2 Sub-70, Dist-3, Dist-3 Sub-70, Dist-4 and Dist-4 Sub-70).


\begin{figure*}[!h]
\centering
\subfloat[]{\label{CMA}\includegraphics[width=0.45\linewidth]{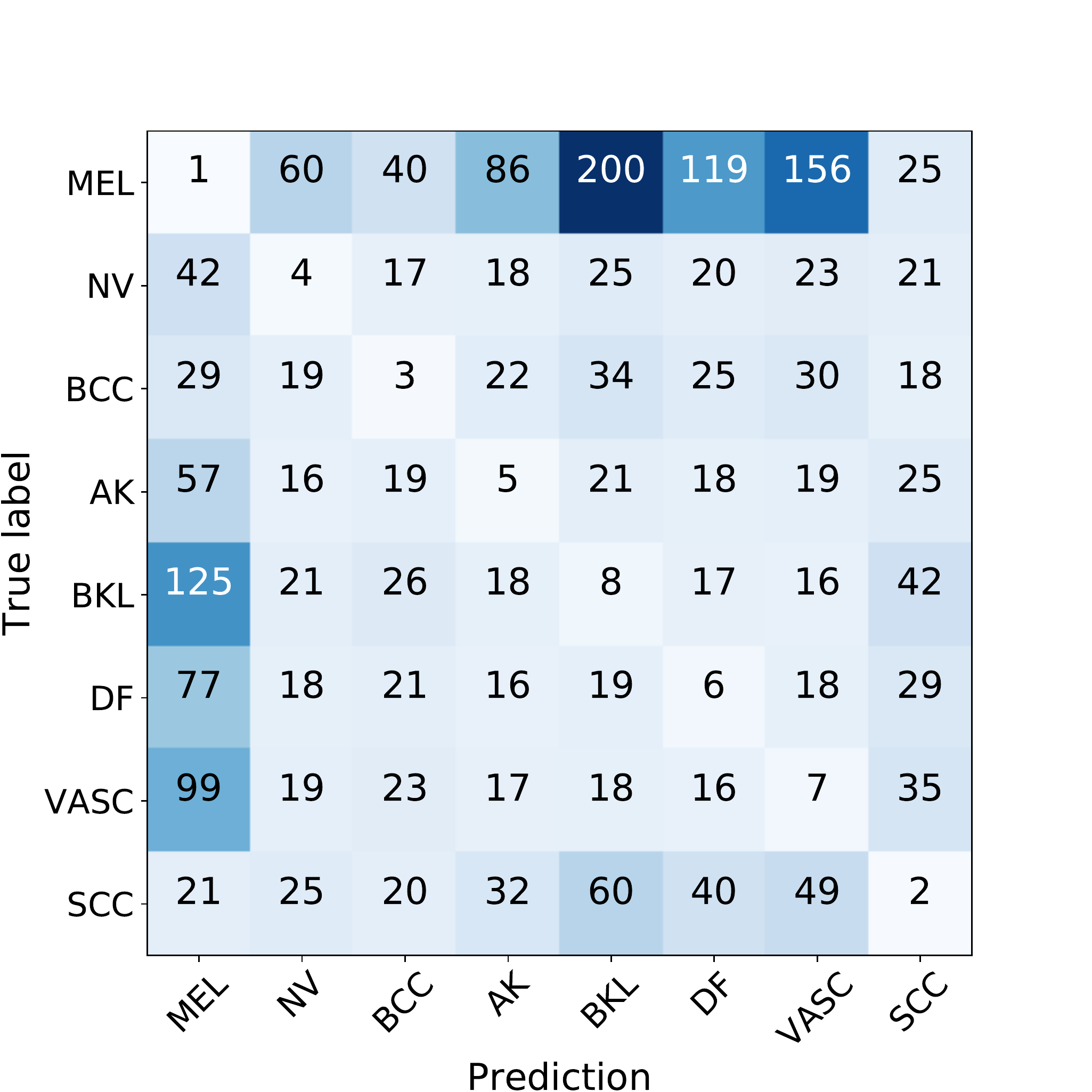}}\hfil
\subfloat[]{\label{CMB}\includegraphics[width=0.45\linewidth]{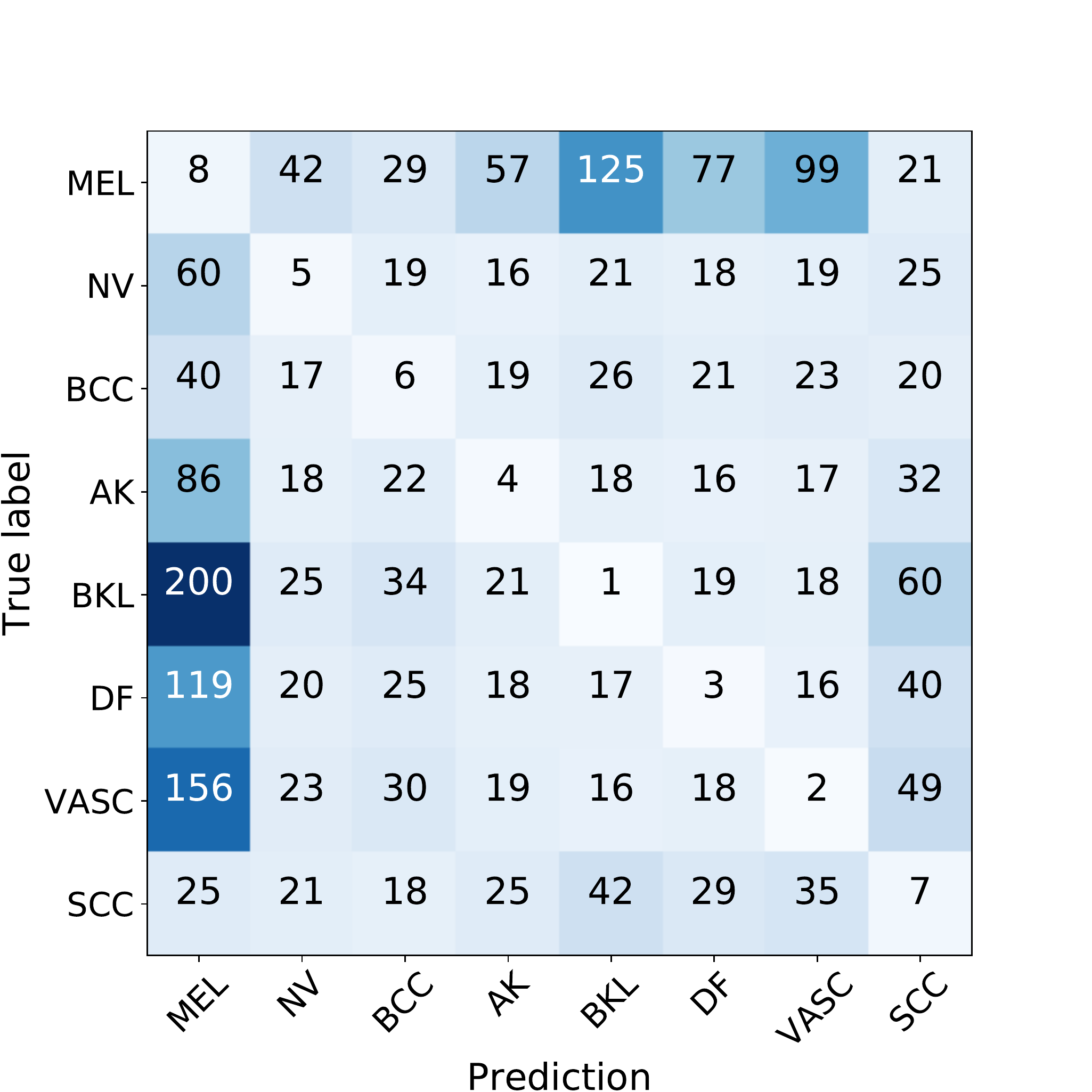}}
\caption{Two examples of cost matrices: (a) Cost Matrix A (emphasizing on cancerous lesions); (b) Cost Matrix B (emphasizing on benign lesions).}\label{costmatrix}
\end{figure*}

\subsection{Base Classifiers Preparation} \label{Experiment_Design}
We prepared the base classifiers to evaluate the fusion approaches performance using 12 different popular CNN architectures (as shown in Table \ref{table1}) with parameters pre-trained on ImageNet \cite{imagenet_cvpr09}. Different CNN architectures have different size of input images:
\begin{itemize}
    \item 331$\times$331: PNASNet-5-Large, NASNet-A-Large.
    \item 320$\times$320: ResNeXt101-32$\times$16d.
    \item 299$\times$299: Xception, Inception-V4, Inception-V3, InceptionResNet-V2.
    \item 224$\times$224: SENet154, Dual Path Net-107, SE-ResneXt101-32$\times$4d, EfficientNet-b7, ResNet152.
\end{itemize}

All the base classifiers were fine-tuned by stochastic gradient decent (SGD) with learning rate $10^{-3}$ and momentum 0.9. The learning rate degraded in 20 epochs by 0.1. We stopped the training process either after 40 epochs or while the validation accuracy was failed to improve for 7 consecutive epochs. Our experiments were implemented using Pytorch, running on a server with 4 GTX 1080Ti 11 GB GPUs. To keep the same batch size 32 in each evaluation, and due to the memory constraint of single GPU, certain CNN architectures were trained with more GPUs:
\begin{itemize}
    \item 2 GPUs: SENet154, EfficientNet-b7, Dual Path Net-107.
    \item 4 GPUs: PNASNet-5-Large, ResNeXt101-32$\times$16d, NASNet-A-Large.
\end{itemize}

Table \ref{table1} illustrates the performance (i.e., accuracy) of all 96 base classifiers on the testing dataset.

\subsection{Design of the Cost Matrix} \label{CM_Design}
As discussed in Section~\ref{Cost_Problem}, it is necessary to have an appropriate cost matrix to enable our ``cost-sensitive'' feature in our active fusion framework. Hence, we create several principles to design an adequate cost matrix:
\begin{itemize}
    \item All the costs should be positive, since it will be item-wise multiplied with the confusion matrix, and we don’t want to get non-positive values in our cost-sensitive confusion matrix.
    \item The cost of the correct predictions should depend on the relative severeness of the corresponding disease. For instance, it should be more valuable (i.e., less cost) to classify a more severe disease (i.e., melanoma) correctly. To figure out the relative severeness relationships among all eight skin lesion classes and better design our cost matrix, we referred to some literature articles regarding to the severeness of those diseases \cite{ CM_2, CM_3, CM_4, CM_5, CM_6, CM_7, CM_9, CM_10, CM_11}. To be simplified and enable the evaluation of our work, based on those references, we heuristically ordered the severeness of the 8 skin lesion classes (from the most severe to the least one) as follows: melanoma (MEL), squamous cell carcinoma (SCC), basal cell carcinoma (BCC), melanocytic nevus (NV), actinic keratosis (AK), dermatofibroma (DF), vascular lesion (VASC), benign keratosis (BKL).
    \item The relative costs of different incorrect predictions should be based on their relative severeness. For instance, misclassifying melanoma (i.e, a deadly cancerous lesion) as benign keratosis should result in much more cost than the opposite scenario.
    \item The cost of correct predictions should be no more than the cost of incorrect predictions.
\end{itemize}

As such, we designed two examples of cost matrices that emphasize on different skin lesion classes (i.e., cancerous lesions vs. benign lesions) to evaluate our work, as illustrated in Figure~\ref{CMA} (i.e., Cost Matrix A) and Figure~\ref{CMB} (i.e., Cost Matrix B).

Let us take the design of Cost Matrix A, that emphasizes on cancerous lesions (i.e., the cost of misclassifying a cancerous lesion is much more than a benign lesion), as an example. Firstly, we assign the cost of correct prediction of each skin lesion class, i.e., $c_{ii}$, $i=1, 2, \dots, m$ (as defined in Section~\ref{Cost_Problem}), according to the relative disease severeness, where predicting a more severe skin lesion class correctly should result in less cost. For instance, we set the cost of correct prediction of MEL (i.e., the most severe one) as 1, and the cost of correct prediction of BKL (i.e., the least severe one) as 8. Secondly, to calculate the relative cost of each incorrect prediction, we follow the equation below:
\begin{equation}
        c_{ij}=\big(\frac{c_{jj}}{c_{ii}}\big)^{2}, \ i \neq j
        \label{misclassifying_cost}
\end{equation}
where as defined in Section~\ref{Cost_Problem}, $c_{ij}$ denote the cost of classifying an instance belonging to class $i$ into class $j$. For instance, if the cost of correct prediction of MEL is 1 and the cost of correct prediction of BKL is 8, the cost of misclassifying an instance belonging to MEL into BKL would be $(\frac{8}{1})^{2}=64$. Last but not least, to ensure the cost of correct predictions are always no more than the cost of incorrect predictions, without loss of generality, we normalized the costs of misclassifications to integers between 16 and 200, using min-max scaling. Figure~\ref{CMA} shows the final result of our designed Cost Matrix A.

To evaluate our framework under different cost matrices, we also designed a Cost Matrix B (as shown in Figure~\ref{CMB}), that emphasizes on benign lesions (i.e., the cost of misclassifying a benign lesion is much more than a cancerous lesion). Cost Matrix B follows the same design steps as Cost Matrix A, other than considering an exactly reverse order of the severeness. For instance, in the design of Cost Matrix B, melanoma became the ``least severe'' one and benign keratosis became the ``most severe'' one.

Our two examples of cost matrices are just for our experimental evaluation purposes. To utilize our framework, the other researchers could always create the cost matrix with their own demands or based on their own best domain expert knowledge.

\begin{figure*}[t]
\centering
\subfloat[]{\includegraphics[width=0.405\linewidth]{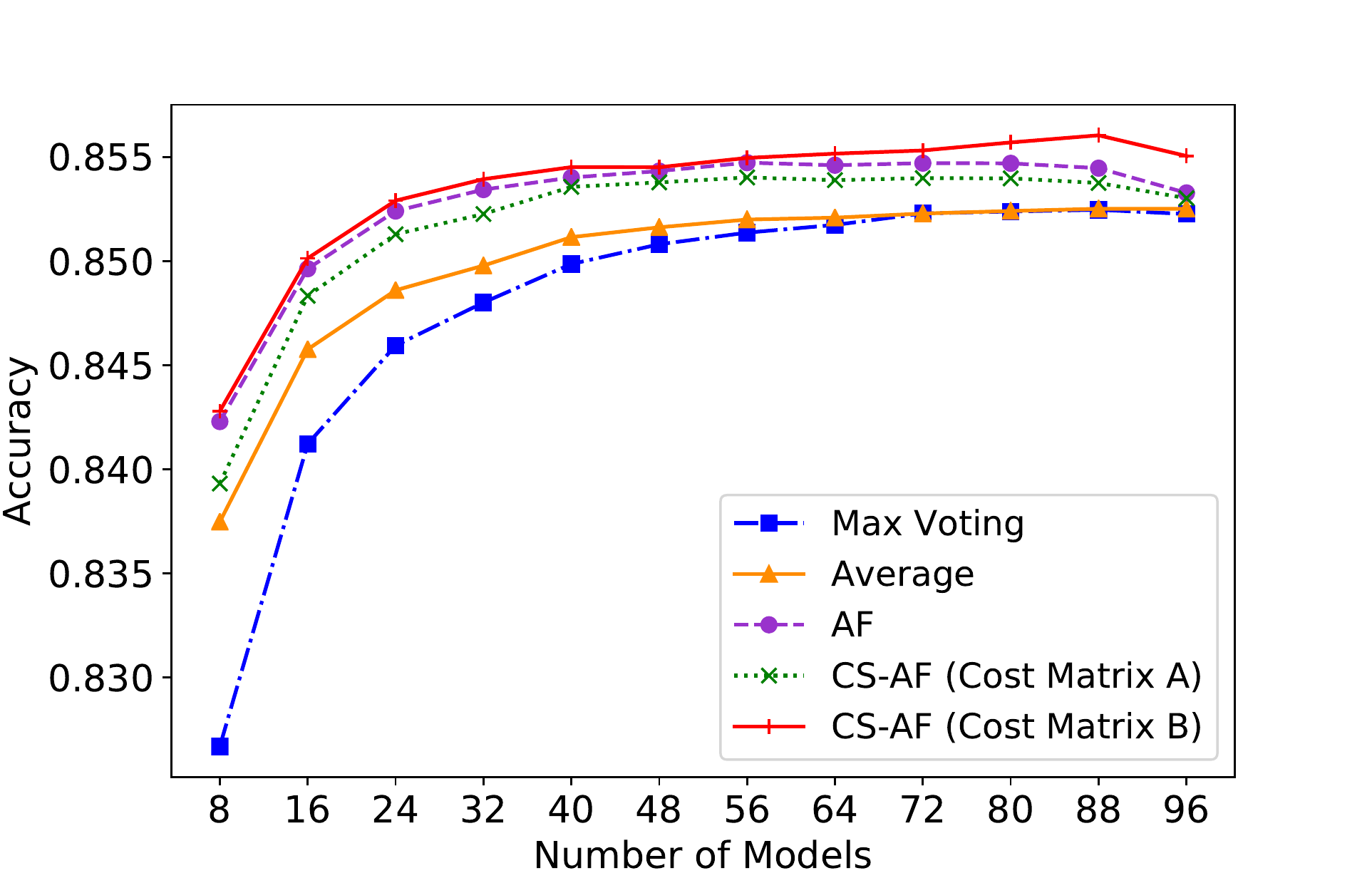}\label{EVA_accuracy}}
\subfloat[]{\includegraphics[width=0.2975\linewidth]{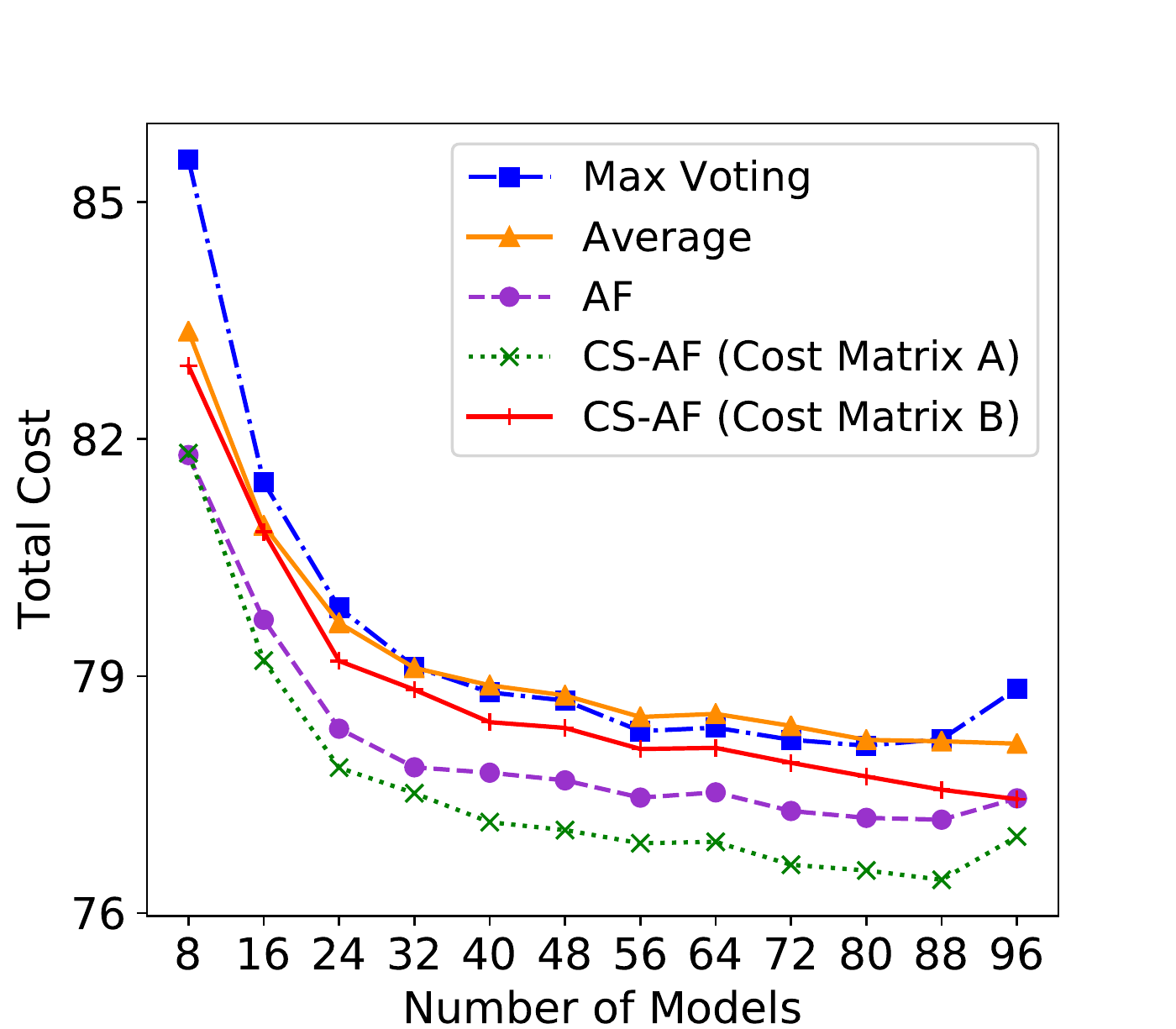}\label{CMA_ACC}}
\subfloat[]{\includegraphics[width=0.2975\linewidth]{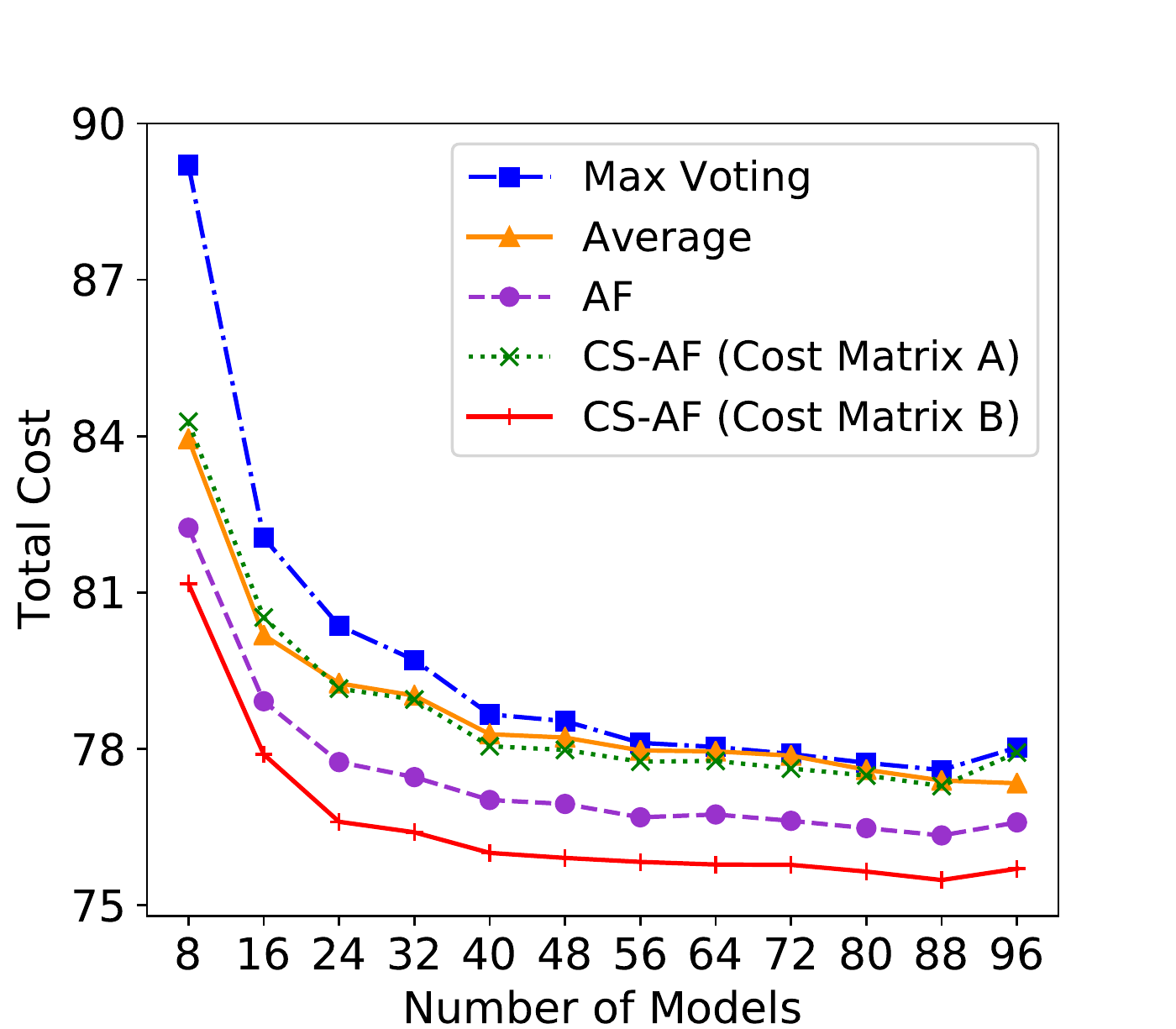}\label{CMB_ACC}}
\caption{Evaluate the Effectiveness of CS-AF. (a) The overall accuracy on our testing dataset; (b) The total cost calculated by Cost Matrix A; (c) The total cost calculated by Cost Matrix B. }\label{accuracy_cost}
\end{figure*}

\subsection{Experimental Procedure}
As described in Section~\ref{Experiment_Design}, we have prepared 96 base classifiers. To evaluate the effectiveness of our active fusion approach extensively, we use different subsets of the 96 base classifiers to form a fusion model. Specifically, given a fusion approach, each time, we randomly select $N$ classifiers among all 96 base classifiers to form a fusion model and evaluate its performance, where $N=8, 16, 24, 32, 40, 48, 56, 64, 72, 80, 88, 96$. For each specific $N$, we will repeat the random selection experiments for 100 times, and use the averaged performance as the final results.

\subsection{Evaluate the Effectiveness of CS-AF}
To demonstrate the effectiveness of our approach, a comparison between two CS-AF implementations (using two different cost matrices to compute the objective weights) and three other fusion approaches has been conducted. The contestants are:
\begin{itemize}
    \item Max Voting Fusion is a static approach, where predictions are combined from multiple base classifiers and only the predicted class with the highest votes will be included in the final prediction.
    \item Average Fusion is another static approach, where it averages the decision vectors of multiple base classifiers and uses the averaged decision vector to make the final prediction.
    \item AF is a baseline active fusion approach by removing the cost-sensitive adjustment step from CS-AF while calculating the objective weights.
    \item CS-AF (Cost Matrix A) is our approach while computing its objective weights using Cost Matrix A.
    \item CS-AF (Cost Matrix B) is our approach while computing its objective weights using Cost Matrix B.
\end{itemize}

Given a contestant fusion approach, we evaluate its effectiveness in terms of (i) its averaged accuracy on our testing dataset (as shown in Figure~\ref{EVA_accuracy}), (ii) its total cost on our testing dataset specified by Cost Matrix A (as shown in Figure~\ref{CMA_ACC}), and (iii) its total cost on our testing dataset specified by Cost Matrix B (as shown in Figure~\ref{CMB_ACC}).
The total cost could be calculated by the sum of the item-wise product between the confusion matrix resulted from our testing dataset and the specified cost matrix. By definition, better fusion approach usually leads to higher accuracy on the testing dataset and lower total cost specified by certain cost matrix. From the results (Figure~\ref{accuracy_cost}), we could observe that:

\begin{itemize}
    \item Compared with the best performed base classifier, ResNeXt101-32$\times$16d, as shown in Table~\ref{table1}, our two implementations of CS-AF and AF consistently achieve over 2\%-5\% higher accuracy on the testing dataset.
    \item For all the fusion approaches, as more base classifiers involved, the accuracy tends to increase and the total cost tends to decrease.
    \item In terms of the accuracy, our two implementations of CS-AF and AF consistently outperform the static fusion approaches (Max Voting and Average), and CS-AF (Cost Matrix B) consistently achieves the best accuracy. Hence, it presents that our proposed active fusion approach is rather effective towards enhancing the overall accuracy.
    \item In terms of the total cost, CS-AF consistently outperforms the other fusion approaches (Max Voting, Average and AF). For instance, while calculating the total cost using Cost Matrix A, CS-AF (Cost Matrix A) always achieves the lowest total cost, and while calculating the total cost using Cost Matrix B, CS-AF (Cost Matrix B) always obtains the lowest total cost. Thus, it demonstrates that our proposed cost-sensitive active fusion approach could adapt to different cost matrices and is optimized to achieve the lowest total cost.
\end{itemize}

\begin{figure*}[!h]
  \centering
  \includegraphics[width=1\linewidth]{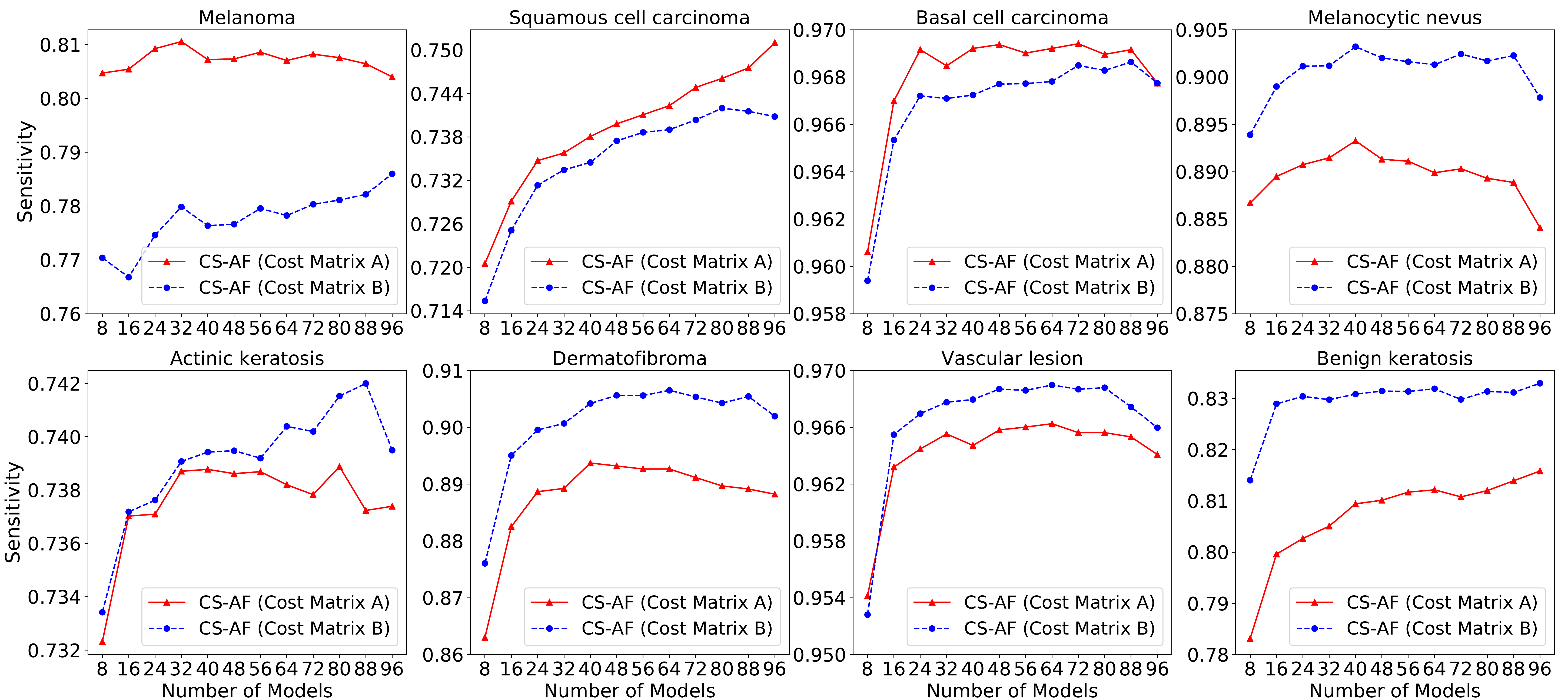}
  \caption{The sensitivity results of each single class of CS-AF (Cost Matrix A) and CS-AF (Cost Matrix B).}
  \label{single_acc}
\end{figure*}

\begin{figure*}[!h]
  \centering
  \includegraphics[width=1\linewidth]{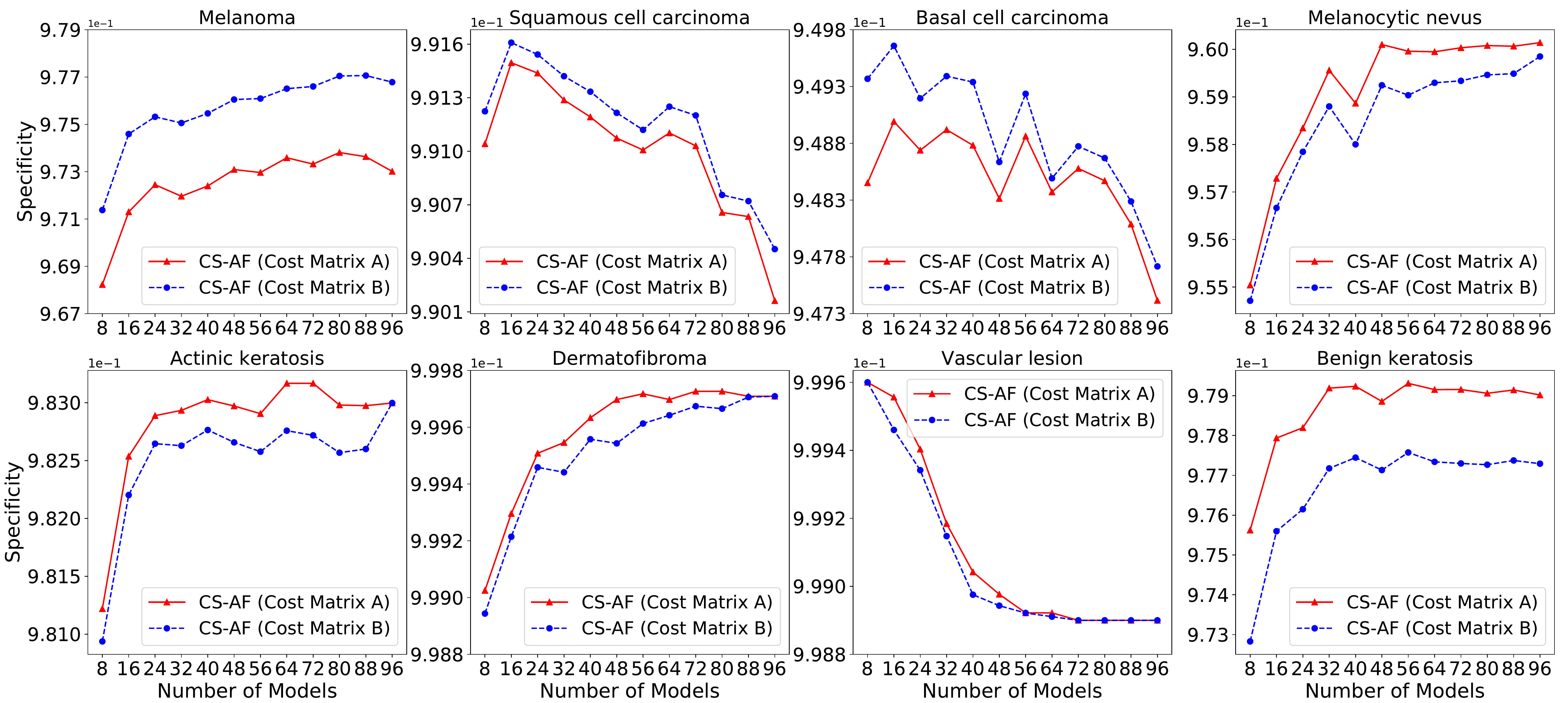}
  \caption{The specificity results of each single class of CS-AF (Cost Matrix A) and CS-AF (Cost Matrix B).}
  \label{specificity}
\end{figure*}

\subsection{Analyze the ``Cost-sensitive'' of CS-AF} \label{effectiveness_costmatrix}
As discussed above, our proposed CS-AF could adapt to different cost matrices and is optimized to achieve the lowest total cost under a specified cost matrix, namely ``cost-sensitive''. In this section, we would like to analyze how such ``cost-sensitive'', while using certain cost matrix, influence the performance of CS-AF on certain single skin lesion classes, thus reducing the total cost. We evaluate single class performances using sensitivity and specificity, defined as follows:

\begin{equation}
        sensitivity=\frac{TP}{TP+FN}
        \label{sensitivity}
\end{equation}
where $TP$ denotes the number of true positives and $FN$ denotes the number of false negatives.

\begin{equation}
        specificity=\frac{TN}{TN+FP}
        \label{specificity}
\end{equation}
where $TN$ denotes the number of true negatives and $FP$ denotes the number of false positives.

Figure~\ref{single_acc} and Figure~\ref{specificity} illustrate the sensitivity and specificity results of each single class of CS-AF (Cost Matrix A) and CS-AF (Cost Matrix B), respectively. We could observe that
\begin{itemize}
    \item Compared with CS-AF (Cost Matrix B), CS-AF (Cost Matrix A) tends to achieve higher sensitivity on more severe cancerous lesion classes (i.e., melanoma, squamous cell carcinoma and basal cell carcinoma), and lower sensitivity on less severe benign lesion classes (i.e., benign keratosis, vascular lesion, dermatofibroma, actinic keratosis and melanocytic nevus)
    \item Compared with CS-AF (Cost Matrix A), CS-AF (Cost Matrix B) tends to achieve higher specificity on those more severe cancerous lesion classes, and lower specificity on those less severe benign lesion classes.
\end{itemize}

As described in Section~\ref{CM_Design}, Cost Matrix A emphasizes on cancerous lesions (i.e., the cost of misclassifyinga cancerous lesion is much more than a benign lesion), while  Cost Matrix B emphasizes on benign lesions (i.e., the cost of misclassifyinga benign lesion is much more than a cancerous lesion). While using Cost Matrix A to compute the objective weights of our CS-AF implementation (i.e., CS-AF (Cost Matrix A)), it tends to increase the TP and FP of cancerous lesion classes and decrease the FN and TN of benign lesion classes, thus resulting in higher sensitivity and lower specificity for cancerous lesion classes. CS-AF (Cost Matrix B) also works in such way accordingly. Therefore, the observations demonstrate that our proposed CS-AF is ``cost-sensitive'', where its performance on certain single skin lesion classes could be flexibly influenced by or adapted to certain well designed cost matrices.

\section{Conclusion} \label{Conclusion}
In this paper, we propose CS-AF, a cost-sensitive multi-classifier active fusion framework for skin lesion classification, where we define two types of weights: the objective weights that are designed according to the classifiers’ reliability to recognize the particular skin lesions, and the subjective weights that are designed according to the relative confidence of the classifiers while recognizing a specific previously “unseen” image (i.e., individuality). We also enable the ``cost-sensitive'' feature in our framework, via incorporating a customizable cost matrix in the design of the objective weights.
In the experimental evaluation, we trained 96 classifiers of 12 CNN architectures as the base classifiers, and compared our CS-AF framework with two static fusion approaches (i.e., Max Voting Fusion and Average Fusion), and a baseline active fusion approach, AF. Our experimental results show that our CS-AF framework consistently outperforms the static fusion competitors in terms of accuracy, and always achieves lower total cost. We also demonstrated our ``cost-sensitive'' feature by using two examples of cost matrices.
In our future work, we plan to
(i) investigate and incorporate other metrics (i.e., other than F1-score) in the design of the objective weights; (ii) design learning-based approach to determine the subjective weights; and (iii) employ and evaluate our CS-AF framework in other medicine-related applications.

\section*{Acknowledgments} Effort sponsored in part by United States Special Operations Command (USSOCOM), under Partnership Intermediary Agreement No. H92222-15-3-0001-01. The U.S. Government is authorized to reproduce and distribute reprints for Government purposes notwithstanding any copyright notation thereon. \footnote{The views and conclusions contained herein are those of the authors and should not be interpreted as necessarily representing the official policies or endorsements, either expressed or implied, of the United States Special Operations Command.}
\bibliography{mybibfile}

\begin{thebibliography}{10}
\expandafter\ifx\csname url\endcsname\relax
  \def\url#1{\texttt{#1}}\fi
\expandafter\ifx\csname urlprefix\endcsname\relax\def\urlprefix{URL }\fi
\expandafter\ifx\csname href\endcsname\relax
  \def\href#1#2{#2} \def\path#1{#1}\fi

\bibitem{zhang2019attention}
J.~Zhang, Y.~Xie, Y.~Xia, C.~Shen, Attention residual learning for skin lesion
  classification, IEEE transactions on medical imaging 38~(9) (2019)
  2092--2103.

\bibitem{gutman2016skin}
D.~Gutman, N.~C. Codella, E.~Celebi, B.~Helba, M.~Marchetti, N.~Mishra,
  A.~Halpern, Skin lesion analysis toward melanoma detection: A challenge at
  the international symposium on biomedical imaging (isbi) 2016, hosted by the
  international skin imaging collaboration (isic), arXiv preprint
  arXiv:1605.01397 (2016).

\bibitem{codella2018skin}
N.~C. Codella, D.~Gutman, M.~E. Celebi, B.~Helba, M.~A. Marchetti, S.~W. Dusza,
  A.~Kalloo, K.~Liopyris, N.~Mishra, H.~Kittler, et~al., Skin lesion analysis
  toward melanoma detection: A challenge at the 2017 international symposium on
  biomedical imaging (isbi), hosted by the international skin imaging
  collaboration (isic), in: 2018 IEEE 15th International Symposium on
  Biomedical Imaging (ISBI 2018), IEEE, 2018, pp. 168--172.

\bibitem{codella2019skin}
N.~Codella, V.~Rotemberg, P.~Tschandl, M.~E. Celebi, S.~Dusza, D.~Gutman,
  B.~Helba, A.~Kalloo, K.~Liopyris, M.~Marchetti, et~al., Skin lesion analysis
  toward melanoma detection 2018: A challenge hosted by the international skin
  imaging collaboration (isic), arXiv preprint arXiv:1902.03368 (2019).

\bibitem{perez2019solo}
F.~Perez, S.~Avila, E.~Valle, Solo or ensemble? choosing a cnn architecture for
  melanoma classification, in: Proceedings of the IEEE Conference on Computer
  Vision and Pattern Recognition Workshops, 2019, pp. 0--0.

\bibitem{di2020saia}
N.~N. Di~Zhuang, K.~Chen, J.~M. Chang, Saia: Split artificial intelligence
  architecture for mobile healthcare systems, arXiv preprint arXiv:2004.12059
  (2020).

\bibitem{tao2014ensemble}
D.~Tao, L.~Jin, Y.~Yuan, Y.~Xue, Ensemble manifold rank preserving for
  acceleration-based human activity recognition, IEEE transactions on neural
  networks and learning systems 27~(6) (2014) 1392--1404.

\bibitem{zhuang2020utility}
D.~Zhuang, J.~M. Chang, Utility-aware privacy-preserving data releasing, arXiv
  preprint arXiv:2005.04369 (2020).

\bibitem{wu2016cost}
P.-Y. Wu, C.-C. Fang, J.~M. Chang, S.-Y. Kung, Cost-effective kernel ridge
  regression implementation for keystroke-based active authentication system,
  IEEE transactions on cybernetics 47~(11) (2016) 3916--3927.

\bibitem{ding2017trunk}
C.~Ding, D.~Tao, Trunk-branch ensemble convolutional neural networks for
  video-based face recognition, IEEE transactions on pattern analysis and
  machine intelligence 40~(4) (2017) 1002--1014.

\bibitem{nguyen2019autogan}
H.~Nguyen, D.~Zhuang, P.-Y. Wu, M.~Chang, Autogan-based dimension reduction for
  privacy preservation, Neurocomputing (2019).

\bibitem{zhuang2017fripal}
D.~Zhuang, S.~Wang, J.~M. Chang, Fripal: Face recognition in privacy
  abstraction layer, in: 2017 IEEE Conference on Dependable and Secure
  Computing, IEEE, 2017, pp. 441--448.

\bibitem{mai2018cluster}
L.~Mai, D.~K. Noh, Cluster ensemble with link-based approach for botnet
  detection, Journal of Network and Systems Management 26~(3) (2018) 616--639.

\bibitem{zhuang2017peerhunter}
D.~Zhuang, J.~M. Chang, Peerhunter: Detecting peer-to-peer botnets through
  community behavior analysis, in: 2017 IEEE Conference on Dependable and
  Secure Computing, IEEE, 2017, pp. 493--500.

\bibitem{zhuang2018enhanced}
D.~Zhuang, J.~M. Chang, Enhanced peerhunter: Detecting peer-to-peer botnets
  through network-flow level community behavior analysis, IEEE Transactions on
  Information Forensics and Security 14~(6) (2018) 1485--1500.

\bibitem{tagarelli2017ensemble}
A.~Tagarelli, A.~Amelio, F.~Gullo, Ensemble-based community detection in
  multilayer networks, Data Mining and Knowledge Discovery 31~(5) (2017)
  1506--1543.

\bibitem{zhuang2019dynamo}
D.~Zhuang, M.~J. Chang, M.~Li, Dynamo: Dynamic community detection by
  incrementally maximizing modularity, IEEE Transactions on Knowledge and Data
  Engineering (2019).

\bibitem{breiman1996bagging}
L.~Breiman, Bagging predictors, Machine learning 24~(2) (1996) 123--140.

\bibitem{schapire1990strength}
R.~E. Schapire, The strength of weak learnability, Machine learning 5~(2)
  (1990) 197--227.

\bibitem{freund1995desicion}
Y.~Freund, R.~E. Schapire, A desicion-theoretic generalization of on-line
  learning and an application to boosting, in: European conference on
  computational learning theory, Springer, 1995, pp. 23--37.

\bibitem{wolpert1992stacked}
D.~H. Wolpert, Stacked generalization, Neural networks 5~(2) (1992) 241--259.

\bibitem{ren2018multi}
F.~Ren, Y.~Li, M.~Hu, Multi-classifier ensemble based on dynamic weights,
  Multimedia Tools and Applications 77~(16) (2018) 21083--21107.

\bibitem{cruz2015meta}
R.~M. Cruz, R.~Sabourin, G.~D. Cavalcanti, T.~I. Ren, Meta-des: A dynamic
  ensemble selection framework using meta-learning, Pattern recognition 48~(5)
  (2015) 1925--1935.

\bibitem{garcia2018dynamic}
S.~Garc{\'\i}a, Z.-L. Zhang, A.~Altalhi, S.~Alshomrani, F.~Herrera, Dynamic
  ensemble selection for multi-class imbalanced datasets, Information Sciences
  445 (2018) 22--37.

\bibitem{marchetti2018results}
M.~A. Marchetti, N.~C. Codella, S.~W. Dusza, D.~A. Gutman, B.~Helba, A.~Kalloo,
  N.~Mishra, C.~Carrera, M.~E. Celebi, J.~L. DeFazio, et~al., Results of the
  2016 international skin imaging collaboration international symposium on
  biomedical imaging challenge: Comparison of the accuracy of computer
  algorithms to dermatologists for the diagnosis of melanoma from dermoscopic
  images, Journal of the American Academy of Dermatology 78~(2) (2018)
  270--277.

\bibitem{bi2017automatic}
L.~Bi, J.~Kim, E.~Ahn, D.~Feng, Automatic skin lesion analysis using
  large-scale dermoscopy images and deep residual networks, arXiv preprint
  arXiv:1703.04197 (2017).

\bibitem{mollineda2007class}
R.~Mollineda, R.~Alejo, J.~Sotoca, The class imbalance problem in pattern
  classification and learning, in: II Congreso Espanol de Inform{\'a}tica (CEDI
  2007). ISBN, 2007, pp. 978--84.

\bibitem{chawla2002smote}
N.~V. Chawla, K.~W. Bowyer, L.~O. Hall, W.~P. Kegelmeyer, Smote: synthetic
  minority over-sampling technique, Journal of artificial intelligence research
  16 (2002) 321--357.

\bibitem{wang2014hybrid}
K.-J. Wang, B.~Makond, K.-H. Chen, K.-M. Wang, A hybrid classifier combining
  smote with pso to estimate 5-year survivability of breast cancer patients,
  Applied Soft Computing 20 (2014) 15--24.

\bibitem{han2005borderline}
H.~Han, W.-Y. Wang, B.-H. Mao, Borderline-smote: a new over-sampling method in
  imbalanced data sets learning, in: International conference on intelligent
  computing, Springer, 2005, pp. 878--887.

\bibitem{bunkhumpornpat2009safe}
C.~Bunkhumpornpat, K.~Sinapiromsaran, C.~Lursinsap, Safe-level-smote:
  Safe-level-synthetic minority over-sampling technique for handling the class
  imbalanced problem, in: Pacific-Asia conference on knowledge discovery and
  data mining, Springer, 2009, pp. 475--482.

\bibitem{maciejewski2011local}
T.~Maciejewski, J.~Stefanowski, Local neighbourhood extension of smote for
  mining imbalanced data, in: 2011 IEEE Symposium on Computational Intelligence
  and Data Mining (CIDM), IEEE, 2011, pp. 104--111.

\bibitem{zhang2016evolutionary}
L.~Zhang, D.~Zhang, Evolutionary cost-sensitive extreme learning machine, IEEE
  transactions on neural networks and learning systems 28~(12) (2016)
  3045--3060.

\bibitem{iranmehr2019cost}
A.~Iranmehr, H.~Masnadi-Shirazi, N.~Vasconcelos, Cost-sensitive support vector
  machines, Neurocomputing 343 (2019) 50--64.

\bibitem{khan2017cost}
S.~H. Khan, M.~Hayat, M.~Bennamoun, F.~A. Sohel, R.~Togneri, Cost-sensitive
  learning of deep feature representations from imbalanced data, IEEE
  transactions on neural networks and learning systems 29~(8) (2017)
  3573--3587.

\bibitem{visa2011confusion}
S.~Visa, B.~Ramsay, A.~L. Ralescu, E.~Van Der~Knaap, Confusion matrix-based
  feature selection., MAICS 710 (2011) 120--127.

\bibitem{van2013macro}
V.~Van~Asch, Macro-and micro-averaged evaluation measures [[basic draft]],
  Belgium: CLiPS 49 (2013).

\bibitem{combalia2019bcn20000}
M.~Combalia, N.~C. Codella, V.~Rotemberg, B.~Helba, V.~Vilaplana, O.~Reiter,
  A.~C. Halpern, S.~Puig, J.~Malvehy, Bcn20000: Dermoscopic lesions in the
  wild, arXiv preprint arXiv:1908.02288 (2019).

\bibitem{tschandl2018ham10000}
P.~Tschandl, C.~Rosendahl, H.~Kittler, The ham10000 dataset, a large collection
  of multi-source dermatoscopic images of common pigmented skin lesions,
  Scientific data 5 (2018) 180161.

\bibitem{liu2018progressive}
C.~Liu, B.~Zoph, M.~Neumann, J.~Shlens, W.~Hua, L.-J. Li, L.~Fei-Fei,
  A.~Yuille, J.~Huang, K.~Murphy, Progressive neural architecture search, in:
  Proceedings of the European Conference on Computer Vision (ECCV), 2018, pp.
  19--34.

\bibitem{zoph2016neural}
B.~Zoph, Q.~V. Le, Neural architecture search with reinforcement learning,
  arXiv preprint arXiv:1611.01578 (2016).

\bibitem{xie2017aggregated}
S.~Xie, R.~Girshick, P.~Doll{\'a}r, Z.~Tu, K.~He, Aggregated residual
  transformations for deep neural networks, in: Proceedings of the IEEE
  conference on computer vision and pattern recognition, 2017, pp. 1492--1500.

\bibitem{hu2018squeeze}
J.~Hu, L.~Shen, G.~Sun, Squeeze-and-excitation networks, in: Proceedings of the
  IEEE conference on computer vision and pattern recognition, 2018, pp.
  7132--7141.

\bibitem{chen2017dual}
Y.~Chen, J.~Li, H.~Xiao, X.~Jin, S.~Yan, J.~Feng, Dual path networks, in:
  Advances in neural information processing systems, 2017, pp. 4467--4475.

\bibitem{chollet2017xception}
F.~Chollet, Xception: Deep learning with depthwise separable convolutions, in:
  Proceedings of the IEEE conference on computer vision and pattern
  recognition, 2017, pp. 1251--1258.

\bibitem{szegedy2017inception}
C.~Szegedy, S.~Ioffe, V.~Vanhoucke, A.~A. Alemi, Inception-v4, inception-resnet
  and the impact of residual connections on learning, in: Thirty-first AAAI
  conference on artificial intelligence, 2017.

\bibitem{he2016deep}
K.~He, X.~Zhang, S.~Ren, J.~Sun, Deep residual learning for image recognition,
  in: Proceedings of the IEEE conference on computer vision and pattern
  recognition, 2016, pp. 770--778.

\bibitem{szegedy2016rethinking}
C.~Szegedy, V.~Vanhoucke, S.~Ioffe, J.~Shlens, Z.~Wojna, Rethinking the
  inception architecture for computer vision, in: Proceedings of the IEEE
  conference on computer vision and pattern recognition, 2016, pp. 2818--2826.

\bibitem{tan2019efficientnet}
M.~Tan, Q.~V. Le, Efficientnet: Rethinking model scaling for convolutional
  neural networks, arXiv preprint arXiv:1905.11946 (2019).

\bibitem{imagenet_cvpr09}
J.~Deng, W.~Dong, R.~Socher, L.-J. Li, K.~Li, L.~Fei-Fei, {ImageNet: A
  Large-Scale Hierarchical Image Database}, in: CVPR09, 2009.

\bibitem{CM_2}
D.-F.~C. Institute, What’s the difference between melanoma and skin cancer?,
  {https://blog.dana-farber.org/insight/2019/12/difference-between-melanoma-and-skin-cancer/}
  (2019 (accessed March 1, 2020)).

\bibitem{CM_3}
H.~Godman, What are the prognosis and survival rates for melanoma by stage?,
  {https://www.healthline.com/health/melanoma-prognosis-and-survival-rates}
  (2017 (accessed March 1, 2020)).

\bibitem{CM_4}
A.~Oakley, Mole, {https://dermnetnz.org/topics/mole/} (2016 (accessed March 1,
  2020)).

\bibitem{CM_5}
A.~A. of~Dermatology~Association., Types of skin cancer,
  {https://www.aad.org/public/diseases/skin-cancer/types/common} (2020
  (accessed March 1, 2020)).

\bibitem{CM_6}
WebMD, Understanding actinic keratosis -- the basics,
  {https://www.webmd.com/skin-problems-and-treatments/understanding-actinic-keratosis-basics}
  (2019 (accessed March 1, 2020)).

\bibitem{CM_7}
M.~C. Staff, Seborrheic keratosis,
  {https://www.mayoclinic.org/diseases-conditions/seborrheic-keratosis/symptoms-causes/syc-20353878}
  (2019 (accessed March 1, 2020)).

\bibitem{CM_9}
N.~Y. U.~L. Health, Surgery for vascular malformations in the torso \& limbs,
  {https://nyulangone.org/conditions/vascular-malformations-in-the-torso-limbs-in-adults/treatments/surgery-for-vascular-malformations-in-the-torso-limbs}
  (2020 (accessed March 1, 2020)).

\bibitem{CM_10}
N.~Y. U.~L. Health, Types of vascular malformations in the torso \& limbs,
  {www.nyulangone.org/conditions/vascular-malformations-in-the-torso-limbs-in-adults/types}
  (2020 (accessed March 1, 2020)).

\bibitem{CM_11}
U.~of~California San Francisco~Health, Basal cell carcinoma and squamous cell
  carcinoma,
  {http://www.ucsfhealth.org/conditions/basal-cell-and-squamous-cell-carcinoma}
  (2020 (accessed March 1, 2020)).

\end{thebibliography}

\end{document}